\documentclass[journal,twoside,web]{ieeecolor}
\usepackage{generic}
\usepackage{cite}
\usepackage{amsmath,amssymb,amsfonts}
\let\labelindent\relax
\usepackage{enumitem}
\usepackage{algorithmic}
\usepackage{graphicx}
\graphicspath{{imgs/}}
\usepackage{pifont}
\usepackage{algorithm,algorithmic}
\usepackage{hyperref}
\hypersetup{hidelinks}
\usepackage{textcomp}
\usepackage{booktabs}
\usepackage{colortbl}
\usepackage{multirow}
\newcommand{\nv}[1]{#1}
\usepackage{tikz}
\usetikzlibrary{positioning,arrows.meta,backgrounds,fit,calc,shapes.geometric}

\providecommand{\refname}{REFERENCES}

\def\BibTeX{{\rm B\kern-.05em{\sc i\kern-.025em b}\kern-.08em
    T\kern-.1667em\lower.7ex\hbox{E}\kern-.125emX}}
\begin{document}
\title{MedClaw: Heuristic Agent Harness for Long-Horizon Surgical Video Reasoning}
\author{Yingying Fan, Penghui Du, Leyan Zhu, Runze He, Zimeng Wu, Yuxuan Zhang, Liang Chen, Jiahao Xie, Jiangtang Wang, Shuai Shao,  Anchao Yang,
Yutong Bai, Yan Wang, IEEE Member
\thanks{This work was supported in part by the National Natural Science Foundation of China under Grant 82402252, in part by the Noncommunicable Chronic Diseases-National Science and Technology Major Project under Grant no.2023ZD0515200. }
\thanks{Yingying Fan is with the School of Automation and Intelligence, Beijing Jiaotong University, Beijing 100044, China, and also with the Department of Electrical and Computer Engineering, University of Maryland, College Park, MD 20742, USA (e-mail: kristewtoday@gmail.com).}
\thanks{Penghui Du, Leyan Zhu, Runze He, Zimeng Wu, Liang Chen, Jiahao Xie are with UniPat.ai. Shuai Shao and Jiangtao Wang are with Suzhou Institute for Advanced Research, University of Science and Technology of China. Yuxuan Zhang is with University of British Columbia. }
\thanks{Anchao Yang and Yutong Bai are with the Department of Neurosurgery, Beijing Tiantan Hospital, Capital Medical University (Email: yang.anchao@163.com; baiyutong@mail.ccmu.edu.cn). Yan Wang is with School of Automation and Intelligence, Beijing Jiaotong University, No.3 Shangyuancun, Haidian District, Beijing, 100044, China. (Email: wangyan9509@gmail.com).}}
\maketitle
\begin{abstract}
Understanding tens-of-minutes surgical videos requires \emph{long-horizon
temporal reasoning}, answering what happens before, after, or across stages of a
procedure by grounding the question in visual evidence spread across time. Existing
approaches handle this poorly: a one-shot vision--language model (VLM) compresses the
whole procedure to fit its context window and loses the detail a
``before'' or ``after'' question depends on, while video agents that \emph{train}
the model where to look are data-hungry and transfer poorly to out-of-domain
surgery. We build an \emph{agent harness} that separates reasoning from
perception and improves by \emph{evolving context rather than optimizing weights}.
A text-only orchestrator plans which evidence to gather and issues an auditable
sequence of tool calls, while frozen vision--language sub-agents execute each call
over the pixels, viewing, cropping, inspecting frames, and retrieving external
knowledge. We further propose a gradient-free, reward-gated \emph{Heuristic Skill
Distillation} loop that mines the agent's own low-scoring traces and keeps a
candidate skill only when it raises a validation reward, yielding reusable
retrieval skills, notably \emph{directed re-look}. Growing an external skill
library rather than tuning weights, the loop adapts from only about 100
labeled examples, far fewer than supervised or reinforcement fine-tuning requires.
To evaluate this agent, we introduce \emph{MedClawBench}, a de-leaked,
doctor-grounded benchmark of $1{,}123$ questions over self-built long neurosurgery
recordings and a held-out public lecture-video test split. Across both datasets
and all four evaluation dimensions, our agent consistently outperforms one-shot
VLMs and general video-agent frameworks, with the largest gains
on the long, out-of-domain neurosurgery videos.
Project page: \url{https://fyycs.github.io/medclaw/}.
\end{abstract}

\begin{IEEEkeywords}
Long-horizon surgical video reasoning,
Agent harness, Heuristic Skill Distillation.
\end{IEEEkeywords}
    
\section{Introduction}
\label{sec:introduction}

\IEEEPARstart{S}{urgical} videos are a primary medium for teaching, credentialing,
and retrospective review~\cite{green2019,birkmeyer}, yet they are among the hardest inputs for medical video
understanding~\cite{kiyasseh,seenivasan2022surgicalvqa,wang2025surgvidlm}. A procedure runs for tens of minutes, its clinically meaningful
information is spread across time, and the answer to ``what is done \emph{before}
or \emph{after} a given step'' depends on how the operation evolves rather than on
any single frame~\cite{wang2025surgvidlm,li2026medscope}. We call this ability, grounding a question about the temporal
evolution of a whole procedure in the visual evidence itself,
\textbf{long-horizon temporal reasoning}. It is precisely the ability a trainee
must acquire from watching operations, and precisely the ability that current
surgical video systems fail to deliver.

\begin{figure}[!htp]
    \centering
    \includegraphics[width=1\linewidth]{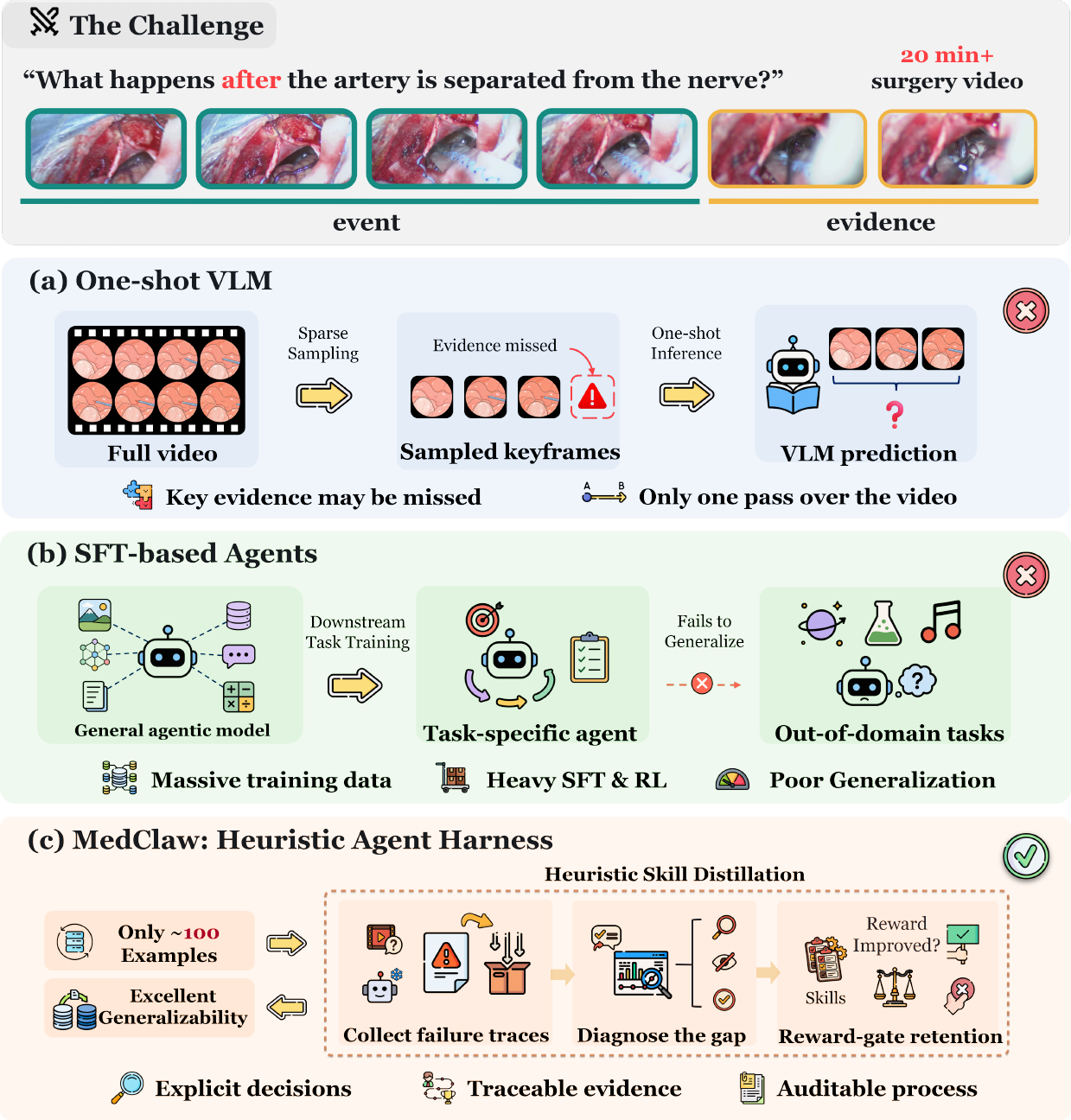}
    \caption{Long-horizon temporal reasoning on a long surgery video,
    where an ``after'' question is answered by evidence that lies outside the
    window depicting the named event. (a) A one-shot VLM reads the whole
    procedure in a single pass, losing the detail the question needs.
    (b) SFT-based agents learn where to look through supervised or reinforcement
    fine-tuning, which needs massive data and transfers poorly out of domain.
    (c) MedClaw keeps every model frozen and instead distills reusable skills
    through Heuristic Skill Distillation, collecting failure traces, diagnosing
    the gap, and retaining a skill only when it passes a reward gate, adapting from
    only about one hundred examples while keeping every decision explicit and
    auditable.}
    \label{fig:intro}
\end{figure}

Two families of methods dominate, and neither fits the task. A \emph{one-shot}
vision--language model (VLM) reads the whole procedure at once~\cite{chen2024longvila}, but a tens-of-minutes
video far exceeds any context window~\cite{qian2024streaming}, so it is compressed to fit and loses the very
detail a ``before'' or ``after'' question needs (Fig.~\ref{fig:intro}). It can only
answer from the coarse, compressed view, and it can never go back to re-examine
\emph{which few seconds actually matter}~\cite{he2025framethinker}. The
task is instead inherently iterative: \emph{decide where to look, look, then
re-decide}, since the answer usually lies \emph{outside} the window that depicts
the named event. This calls for an agent, and a second family builds one, but
\emph{trains} the model where to look through supervised fine-tuning (SFT)
or reinforcement learning (RL)~\cite{zhang2025rewatch,yang2026longvt}. That demands large amounts of task-specific trajectory
data, overfits the training distribution, and transfers poorly to an out-of-domain,
long-horizon setting such as long neurosurgery, precisely the regime where the
ability is most needed and labeled data is scarcest~\cite{ward2021challenges,demir2023deep}.

We therefore ask whether the ability to decide where to look can be obtained
in a \emph{training-free} manner, and our central idea is to
\textbf{improve by evolving context rather than optimizing weights}. We separate the two abilities the task
needs, \emph{deciding where to look} and \emph{actually looking}, and hand each to
a frozen model coordinated only in context. Our
\textbf{agent harness} pairs a
text-only \textbf{orchestrator}, which reasons about the question and issues an
auditable sequence of tool calls, with \textbf{frozen vision--language sub-agents}
that execute those calls on the pixels: viewing the full video, cropping a
candidate time window (\texttt{crop\_video}), inspecting individual frames
(\texttt{select\_frame}, \texttt{crop\_frame}), and querying a surgical knowledge
base. Localization follows a \emph{coarse-to-fine} strategy, moving from the whole
procedure down to the exact moment. \emph{No weight is ever updated}: the
orchestrator supplies strategy, the sub-agents supply perception, and everything
happens in context, which turns the otherwise latent ``which segment to look at''
into a inspectable, re-examinable action trace.

Which retrieval strategies matter, and how are they reused without any parameter
update? Open-ended surgical QA has \emph{no} automatic success signal, so
self-evolving loops that keep whatever the model proposes accumulate
plausible-but-useless rules that can even lower answer quality. Our \textbf{Heuristic
Skill Distillation} (HSD) loop is instead \emph{gradient-free and reward-gated}: it reads the
agent's own low-scoring traces, diagnoses the missing temporal evidence, proposes
candidate strategies, and retains one \emph{only} when it lifts a validation reward
scored on four dimensions (correctness, detail, context, temporal understanding),
so a strategy is kept because it \emph{works}, not because it looks plausible (Fig.~\ref{fig:intro}). Each
retained strategy, notably \textbf{directed re-look} (revisiting the window
adjacent to a localized event), is stored as a \textbf{skill} that names when it
applies and how to call the existing tools, and the orchestrator holds it in a
\emph{skill library} and consults it when a question matches its trigger. Since the loop edits context, it is strikingly \emph{data-efficient}: it
mines its skills from about \emph{one hundred} preference examples, where the SFT
and RL agents we compare against fit millions of parameters to far larger labeled
sets and transfer poorly beyond them.

Evaluating this agent requires a benchmark that truly isolates long-horizon
temporal reasoning. However, existing resources do not: they suffer from four
\emph{shortcuts}, questions with no long-horizon structure (short clips, or long
videos that still ask only single-moment questions), a narrator who speaks the
answer aloud, a ``standard next step'' that follows from textbook priors, and
questions answerable from a single frame. We therefore introduce \textbf{MedClawBench}, a
\emph{de-leaked}, doctor-grounded benchmark of $1{,}123$ questions: $259$ over
\emph{long} neurosurgery recordings (microvascular decompression, tumour
resection, epileptogenic-lesion resection) and $864$ from the held-out test split
of the public SVU-31K corpus~\cite{wang2025surgvidlm}. We validate the automatic
judge against expert ratings on both splits ($\kappa$) so reported scores reflect
clinical judgment rather than a self-referential metric.

Our contributions are as follows:
\begin{itemize}
\item A reproducible \textbf{agent harness} that factorizes
long-video reasoning into a text \textbf{orchestrator} and \textbf{frozen
vision--language sub-agents}. The orchestrator gathers evidence through a small set
of tools, viewing the full video, cropping a candidate time window
(\texttt{crop\_video}), inspecting individual frames (\texttt{select\_frame},
\texttt{crop\_frame}), and querying a surgical knowledge base, so evidence seeking
becomes an auditable trace of tool calls while every model weight remains frozen.
\item A \textbf{Heuristic Skill Distillation} loop that improves the agent by evolving
context rather than optimizing weights: from roughly one hundred preference
examples it discovers reusable temporal evidence-retrieval skills, notably
directed re-look, under objective reward gating, and packages each as a
\textbf{skill} in a separate library
that the orchestrator consults when a question matches the skill's declared
trigger, improving the agent with no parameter update and keeping every retrieval
decision auditable.
\item \textbf{MedClawBench}, a de-leaked, doctor-grounded benchmark
($259$ self-built neurosurgery questions and $864$ held-out SVU questions,
$1{,}123$ in total, blind-filtered and $\kappa$-validated) that isolates
long-horizon temporal reasoning on surgical videos.
\item \textbf{State-of-the-art results.} Our agent consistently outperforms
one-shot VLMs and general video-agent frameworks across both
datasets and all four evaluation dimensions, with the largest margins on the
long, out-of-domain neurosurgery videos where long-horizon temporal reasoning
matters most.
\end{itemize}

\section{Related Work}
\label{sec:relatedwork}

\subsection{Agent Harnesses for Video Reasoning}
A growing line of work casts video understanding as tool-augmented reasoning, where a
language model reasons and acts in an interleaved loop~\cite{yao2022react} and calls
external tools~\cite{qin2024toolllm,patil2024gorilla,hao2023toolkengpt}. For long
videos, these systems adaptively retrieve evidence, selecting or sampling informative
frames~\cite{gao2023mist,yao2025k,yao2025generative,jeoung2024adaptive} and localizing temporal windows via multi-turn spotlighting or agentic synthesis~\cite{he2025framethinker,ge2025framemind,liu2026longvideoagent}, 
often using reinforcement learning to shape retrieval or broader spatiotemporal reasoning policies~\cite{feng2026video,li2025videochat,liu2025visual,chen2026scaling,wang2026videorft,yang2026longvt}.
A convenient way to organize these behaviors is an \emph{agent harness} that
separates a planning text orchestrator from frozen vision--language sub-agents, with
all coordination carried in context~\cite{zhang2026rewardharness}, and such harnesses
also serve as auditable evaluators or task solvers on realistic
benchmarks~\cite{zhang2026clawbench,zhuge2024agent}. In the clinical domain,
MedScope~\cite{li2026medscope} interleaves reasoning with coarse-to-fine tool calls to
``think with videos,'' and a broader line of medical agents trains tool-augmented
reasoning with reinforcement
learning~\cite{jiang2025incentivizing,jiang2026ibisagent}. These methods, however,
train the backbone itself to decide where to look, which demands large task-specific
trajectory data and transfers poorly to an out-of-domain, long-horizon setting such as
long neurosurgery. In contrast, we keep both the orchestrator and the sub-agents
frozen and draw retrieval strategies from an external skill library, so the gain is
attributable to knowing how to look rather than to a fine-tuned backbone. To our
knowledge, this is the \emph{first} training-free agent harness for long-horizon temporal
reasoning on surgical video.

\subsection{Heuristic Learning beyond Gradients}
A complementary paradigm improves a frozen model by searching over its \emph{context}
rather than its \emph{weights}. Reasoning-bootstrapping and self-feedback methods let a
model rewrite its own reasoning or outputs from verbal or reward
signals~\cite{madaan2023self,shinn2023reflexion},
experiential agents distill successful trajectories into reusable skills or memories
recalled later~\cite{wang2023voyager,zhao2024expel,liu2026simplemem,xia2025agent0}, and
context- and prompt-optimization methods edit instructions or context to raise an
objective without back-propagation~\cite{yang2024large,khattab2023dspy,zhang2025agentic},
with reward-gated variants keeping only edits that pass an explicit
check~\cite{zhang2026rewardharness}. Many of these presuppose an environment that scores
success automatically. Without one, as in open-ended surgical question answering,
plausible-but-useless skills can accumulate on the model's own judgment. We instantiate
this heuristic-learning idea as \textbf{Heuristic Skill Distillation}: the agent
diagnoses its own blind failures against the gold, distills only operational lessons
into candidate skills, and keeps a candidate only when it raises a validation
four-dimensional reward. While such context optimization has been explored mainly for
text and general multimodal tasks, we bring it to surgical video, distilling temporal
evidence-retrieval skills for a frozen clinical-video agent.
\subsection{Long Surgical and Clinical Video Understanding}
Surgical procedures are long, so reasoning over them is fundamentally a
long-video problem. Early work focused on localized, clip-level recognition of phases,
tools, and actions~\cite{kiyasseh,surgtoolloc2023}, and early surgical VQA
answered questions about a single endoscopic
scene~\cite{seenivasan2022surgicalvqa}, later grounding the answer in a
domain-specialized language model~\cite{he2024pitvqa}. To move beyond short clips,
recent efforts scale to full-length procedures: SurgVidLM contributes the large-scale
SVU-31K corpus of surgical video--instruction pairs~\cite{wang2025surgvidlm}, from whose
held-out test split we draw the lecture-video portion of our benchmark. OphClip
pretrains an ophthalmic surgical video--language model~\cite{hu2025ophclip}, 
MedGen scales granularly annotated medical video~\cite{wang2025medgen}, 
and corpora such as
SurgPub-Video~\cite{li2025surgpub}, SurgBench~\cite{zhao2025surgbench},
SurgViVQA~\cite{surgvivqa2025}, and EyePCR~\cite{eyepcr2025} broaden surgical VQA
coverage. General long-form video understanding has advanced through memory-augmented
and long-context models~\cite{qian2024streaming,he2024malmm,li2024llama,chen2024longvila,ma2025drvideo,kwan2026video}
and strong open foundation VLMs~\cite{wu2025qwen,zhu2025internvl3,zhang2025videollama}
surveyed in~\cite{liang2026comprehensive,tang2025video}, but these are rarely evaluated
on long surgical procedures. Medical vision--language agents likewise equip a model
with clinical tools for diagnosis and
interpretation~\cite{li2024mmedagent,fathi2025aura,wang2025medagent,wang2026smr}.
Despite this progress, most medical video QA resources use short, pre-trimmed clips with
temporally local content, and several carry spoken narration that states the answer
aloud. Fine-grained, evidence-annotated clinical data~\cite{li2026medscope} begins to
encourage grounding, yet the long-horizon setting, aggregating evidence for what happens
before or after a step across a full-length procedure, remains largely unaddressed. Our
benchmark isolates exactly this ability, pairing long neurosurgery recordings with
a held-out lecture-video split under a blind filter that discards questions answerable
without watching the video.

\section{Method}
\label{sec:method}

\begin{figure*}[!htp]
    \centering
    \includegraphics[width=1\linewidth]{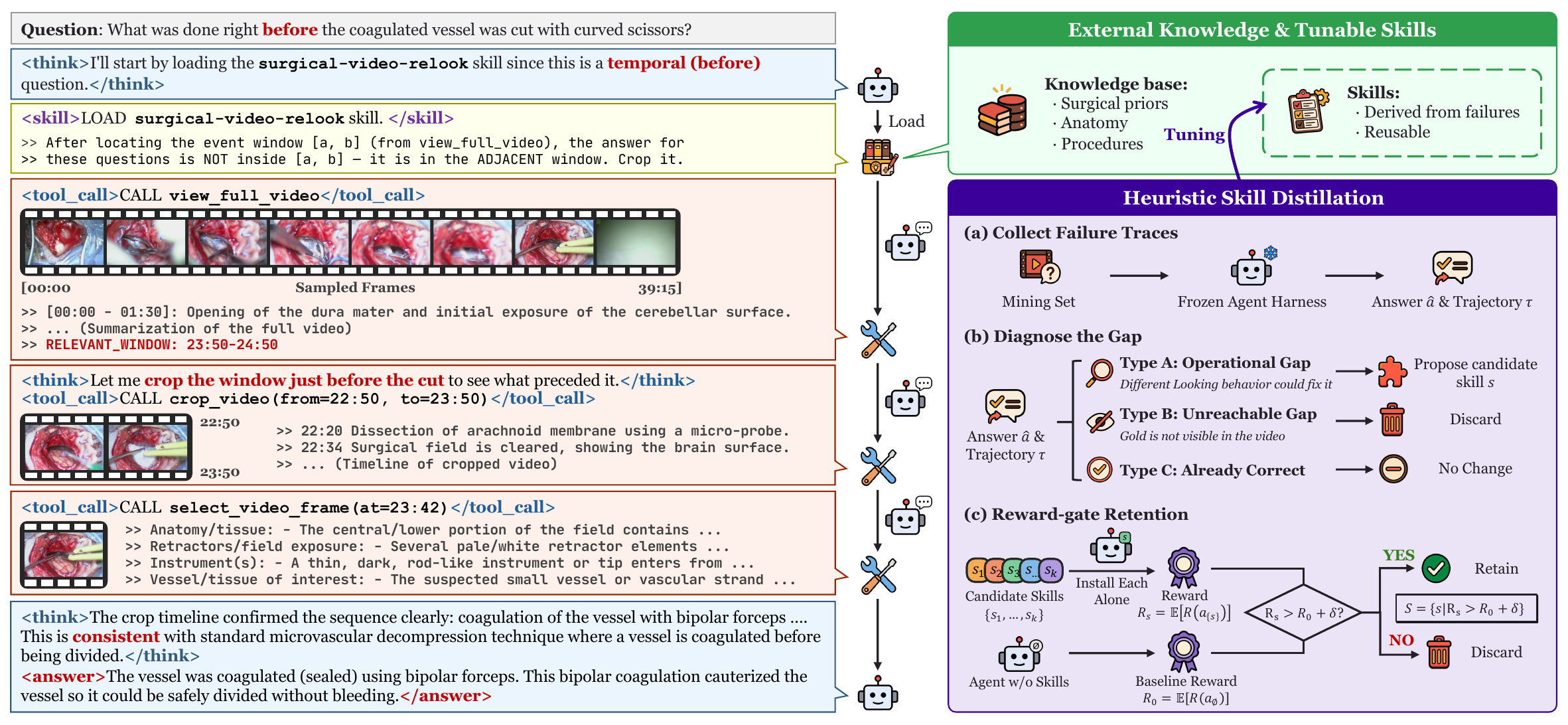}
    \caption{\textbf{Overview of MedClaw.} \emph{Left:} a real trace on a temporal
    ``before'' question, where the orchestrator loads the relevant skill and issues
    an auditable sequence of tool calls that frozen vision sub-agents execute on the
    pixels, viewing the full video to localize a window, cropping it, and inspecting
    a frame, before grounding its answer.
    \emph{Top right:} the orchestrator is backed by an external knowledge base queried via RAG and by a library of
    tunable skills derived from past failures. \emph{Bottom right:} these skills come
    from our \emph{Heuristic Skill Distillation} loop, which collects
    low-scoring traces on a mining set, diagnoses each gap as operational, unreachable, or already correct, and retains a candidate only if it raises the four-dimensional reward
    on a validation set.}
    \label{fig:main}
\end{figure*}

\subsection{Task Formulation}
\label{sec:task}
Given a surgical video $V$ and a natural-language question $q$, the
task is to produce a free-form answer $a$ grounded in what the video actually
shows. We deliberately keep answers open-ended rather than multiple-choice: a
distractor-based format lets a model exploit textual priors~\cite{mmreason}, whereas a free-form
answer must name the specific instruments, structures, and actions observed in
\emph{this} procedure. 
To probe how an agent handles both the \emph{visual-detail} and the
\emph{temporal} sides of long-horizon surgical video understanding, we follow
SVU-31K~\cite{wang2025surgvidlm} and group every question into two reasoning types.
\textbf{Visual Perception Reasoning} (VPR) questions, once the right moment is
localized, turn on a \emph{visual detail} at that moment, such as which instrument
is in use, an anatomical structure, a count, or a finding. \textbf{Visual Temporal Reasoning}
(VTR) questions instead turn on the \emph{process} itself, how the procedure
unfolds: what happens \emph{after} a given step, what was done \emph{before} or in
preparation for an event, or how a state changes over a long horizon. For VTR the
evidence often lies \emph{outside} the window that depicts the named event, so it
can be recovered only by relating moments across the procedure, which makes VTR the
more temporally demanding of the two.


\subsection{Overview}
\label{sec:overview}
Answering a long-horizon temporal question in one forward pass is hard for a
fixed model since the evidence sits in a few seconds out of tens of minutes,
requiring the system to decide where to look, inspect that evidence, and revise its decision. We build an
\emph{agent harness} with three parts that each do only what they are good at.
A text \textbf{orchestrator} $\pi_\theta$ plans which evidence to gather and when to
stop (Section~\ref{sec:orchestration}); a \textbf{tool set} $\mathcal{T}$ exposes
the video at three granularities and adds a retrieval tool over an external
knowledge base (Section~\ref{sec:tools}); and frozen vision \textbf{sub-agents}
$\psi$ answer the concrete visual query each tool is handed
(Section~\ref{sec:subagent}). No weights are updated: neither $\pi_\theta$ nor
$\psi$ is fine-tuned, and competence comes from the loop and its skill library
rather than from updated parameters. On top of this
frozen harness, a \emph{Heuristic Skill Distillation} loop
(Section~\ref{sec:heuristic}) mines reusable retrieval skills from the harness's own
low-scoring trajectories and stores them in a tunable skill library. Fig.~\ref{fig:main} provides an overview of the framework: the left panel illustrates an end to end inference trajectory, where the orchestrator loads
a relevant skill and issues tool calls executed by the vision sub-agents, while the panel
right shows the external knowledge base, the skill library, and the distillation
loop that fills it.

\subsection{Orchestration}
\label{sec:orchestration}
Let $V$ be a surgical video and $q$ a question. The orchestrator is a
text-only policy $\pi_\theta$ (Opus-4.8) with \emph{frozen} parameters $\theta$
that never sees $V$ directly, and it acts through a fixed tool set
$\mathcal{T}=\{T_1,\dots,T_5\}$ (Section~\ref{sec:tools}), each tool being a call
$T_i(V,\cdot)$ answered by the frozen vision sub-agents $\psi$
(Section~\ref{sec:subagent}). Following a reason-act loop~\cite{yao2022react},
at step $t$ the orchestrator conditions on the running context $c_t$ to emit an
action $a_t$, and the selected tool then returns a textual observation $o_t$:
\begin{equation}
a_t=(T_{i_t},u_t)\sim \pi_\theta(\cdot \mid c_t),
\qquad o_t = T_{i_t}(V,u_t),
\end{equation}
where $u_t$ are the corresponding tool arguments (e.g.\ a time range). The context is then extended as
$c_{t+1}=c_t \oplus (a_t,o_t)$, and the loop repeats until the policy emits a
$\textsc{stop}$ action and produces the final answer
$a=\pi_\theta(c_{T})$. All state lives in $c_t$, so no gradient step or external
memory is involved, and the entire trajectory
$\rho=\big((a_1,o_1),\dots,(a_T,o_T)\big)$ is recorded. Because the orchestrator
has no direct access to $V$, every visual need must be externalized as an explicit tool call,
which makes $\rho$ fully auditable and allows gains to be attributed to specific retrieval behaviors.

\subsection{Tool Suite}
\label{sec:tools}
The tool set $\mathcal{T}$ forms a coarse-to-fine progression from the whole clip
to a segment to a single frame, so the orchestrator localizes first and then
verifies the exact instrument, structure, or action at the moment that matters.
Each tool takes the video and typed arguments and returns a textual observation.

\begin{itemize}[leftmargin=1.2em,itemsep=2pt]
\item \texttt{view\_full\_video}$(V,q)$ localizes \emph{where} to look. It runs a
\emph{coarse-to-fine} pass: a Stage-1 low-resolution reading of the full
procedure emits a dense timeline, one short description per time interval
$[a_k,b_k]$ of what is visible in that interval, together with a candidate time
window for the queried event. A Stage-2 re-watch of that widened window at higher
resolution then returns the tightest relevant sub-window $[s^\star,e^\star]$ that
actually contains the queried event. This puts temporal localization on the model that can
see the pixels rather than on the text orchestrator.
\item \texttt{crop\_video}$(V,s,e,q)$ inspects \emph{what} happens in a chosen
range $[s,e]$. It re-reads that segment at crop resolution and returns a dense
description, and is the tool the orchestrator uses to read the localized window
or an adjacent one.
\item \texttt{select\_video\_frame}$(V,t,q)$ returns a description of the single
frame at timestamp $t$ from an independent frame model, used as a cross-model
check of the specific instrument or action at that instant.
\item \texttt{crop\_video\_frame}$(V,t,x_1,y_1,x_2,y_2,q)$ extracts the frame at
$t$, crops the normalized spatial region $(x_1,y_1,x_2,y_2)$, and describes that
region, for fine spatial disambiguation of small or overlapping structures.
\item \texttt{search\_surgical\_kb}$(q,\text{type},k)$ is a
retrieval-augmented-generation (RAG) tool that returns the top-$k$ passages from a
surgical knowledge base when a term needs grounding. The knowledge base holds
general surgical background only, with no benchmark answers: PubMed abstracts
retrieved for surgical keywords, and literature-grounded summaries of surgical
techniques and endoscopic visual cues generated from those keywords, embedded with
a sentence encoder and indexed in FAISS under cosine similarity.
\end{itemize}

\subsection{Vision Sub-agents}
\label{sec:subagent}
Every tool call is executed by frozen vision sub-agents $\psi$ that perform actual perception and return text. They employ no planning or task-specific weights, and comprise two experts that map directly onto the two
granularities of the tool set.
\begin{itemize}[leftmargin=1.2em,itemsep=1pt,topsep=2pt]
\item A \textbf{video expert} (Gemini-3.1-flash-lite/Gemini-3.5-flash) serves
\texttt{view\_full\_video} and \texttt{crop\_video}: it ingests a clip or a
temporal segment and returns timelines, per-interval descriptions, and localized
windows, handling everything that requires reasoning over time.
\item A \textbf{frame verifier} (GPT-5.5) serves
\texttt{select\_video\_frame} and \texttt{crop\_video\_frame}: it ingests a single
frame or a spatial crop of one frame and returns a description of that instant,
giving an independent, higher-resolution check on a specific structure.
\end{itemize}
Keeping perception behind this fixed interface makes $\psi$ interchangeable, so the
same orchestrator $\pi_\theta$ runs over a cheaper or stronger backbone with no
other change, which we exploit to separate the contribution of the harness from
that of the backbone.

\subsection{Heuristic Skill Distillation}
\label{sec:heuristic}
The harness above is fixed, but which retrieval behaviours it should prefer is not
obvious a priori. We therefore add a heuristic skill distillation loop
(Fig.~\ref{fig:main}, bottom right) that discovers reusable retrieval strategies
from the harness's own mistakes and stores them in an external skill library
$\mathcal{S}$ the orchestrator consults at inference. Learning here
evolves this library rather than any model weight, and it is strikingly
data-efficient: it uses only $100$ labeled questions from the SVU training split,
kept disjoint from every evaluation set and split once into a $60$-question
\emph{mining set} on which skills are distilled and a $40$-question \emph{validation
set} on which they are gated, whereas the trained agents we compare against fit
millions of parameters to far larger corpora.

\noindent\textbf{Mining candidate skills from failed traces.}
We walk the mining set one question at a time. On each question, the inference-time
orchestrator first answers \emph{blind}: without seeing the gold, it uses the same
tools as at test time (a \texttt{view\_full\_video} pass, then \texttt{crop\_video}
and \texttt{select\_video\_frame} as needed) and produces an answer $\hat{a}$ and
its trajectory $\rho$. It is then shown the gold, asked to
diagnose the gap between $\hat{a}$ and the gold, and to classify it into one of
three cases. A type~A gap is \emph{operational}: the gold's content was present in
the video and a different way of looking would have caught it, by cropping a
tighter or adjacent window, escalating the crop resolution, selecting a different
frame, or disambiguating two look-alike instruments or structures. A type~B gap is
\emph{unreachable}: the gold refers to something no amount of looking can recover,
such as off-screen narration, the surgeon's stated intent, or a pathology result,
or the backbone simply cannot resolve the pixels. Type~C means the blind answer was
already correct. Only type~A carries an actionable lesson. When the orchestrator
finds one, it writes the lesson itself through the native \texttt{skill\_manage}
tool: it first inspects the current library and, if a skill already states this
same rule, updates that entry (sharpening it or adding the new example) rather than
creating a near-duplicate, otherwise it adds a new skill. Walking the mining set
this way accumulates a small set of candidate skills, capped at five, each mined
directly from a trajectory that failed rather than supplied by us. 

\noindent\textbf{Reward-gated retention.}
Mining only proposes candidates, and whether keeping each one is decided by the
validation reward. Let $R(a)\in[1,5]$ be the mean of four judge dimensions per answer. Before mining, we measure the
empty-library reward $R_0=\mathbb{E}[R(a_{\emptyset})]$, the no-skill baseline, on the validation set. We then test mined candidates one at a time. To
test a candidate $s$ we install it alone, re-answer the entire validation set, re-judge,
and take $R_s=\mathbb{E}[R(a_{\{s\}})]$. The candidate is kept only
if it clears both gates: it must raise the mean four-dimensional reward by a margin,
$R_s>R_0+\delta$ with $\delta=0.05$ on the $1$--$5$ scale, \emph{and} it must not
drop any individual validation question by two points or more. A rejected skill is removed and leaves no trace. This is what
separates the loop from reward-based post-training: the reward gates an auditable,
perception-level skill in an external library, and never edits a weight or a fixed
prompt.



\noindent\textbf{The learned skill library.}
Each retained skill names the condition under which it fires and how to call the
tools with concrete arguments, so its trigger is precise and every retrieval it
prompts stays visible in the trajectory. In practice, the loop converges to a
compact library dominated by one temporal skill, \emph{directed re-look}. The rule
it mines is that for a question about what precedes or follows a located event, the
answer lies in the window \emph{adjacent} to that event rather than inside it. Once
the event window is localized, the skill therefore crops and re-reads the following
window for an ``after'' question, or the preceding window for a ``before'' or
preparation question, and grounds the answer there rather than in the event window
itself. This is exactly the failure mode that our VTR partition is built to stress,
and the harness recovers it not by design but by mining and reward-gating its own
temporal mistakes.

\section{MedClawBench}
\label{sec:benchmark}

We introduce \textbf{MedClawBench}, a benchmark for open-ended question answering
over \emph{long} surgical videos (an $18.4$-minute median, with $41\%$ running beyond
twenty minutes; Fig.~\ref{fig:benchmark_stats}(b)), built to measure
\emph{long-horizon temporal reasoning} in the
medical domain. By construction, every question is free of the four
shortcuts that let existing resources be answered without long-horizon reasoning:
weak temporal structure, narration leakage, textbook-prior answerability, and
single-frame answerability.
Table~\ref{tab:dataset_compare} positions it against existing surgical video QA
datasets: frame-level resources such as
EndoVis18-VQA~\cite{allan20202018} and PitVQA~\cite{he2024pitvqa}, and the
few-frame clip-level SurgViVQA~\cite{surgvivqa2025} carry no long-horizon
structure, though as pure operative footage their questions cannot be answered
without the video. The only full-video corpus that also asks temporal questions,
SVU-31K (test)~\cite{wang2025surgvidlm}, is built on much shorter videos ($32\%$ under
ten minutes) and, being narrated lecture material, leaks answers through speech
yet applies no de-leaking. MedClawBench is the only entry that combines long
surgical videos, explicit de-leaking, and
neurosurgery content in a single benchmark.

\begin{table}[!t]
\centering
\caption{Comparison with surgical-video QA datasets.
\textbf{Video len.}: median source-video duration.
\textbf{$\ge 20$\,min}: share of videos at least twenty minutes long.
\textbf{Temp.}: long-horizon temporal reasoning.
\textbf{De-leak}: not answerable without watching the video (by explicit filtering or by the nature of the source).
\textbf{Neuro.}: includes neurosurgery videos.
Durations and percentages are rounded.}
\label{tab:dataset_compare}
\setlength{\tabcolsep}{4pt}
\renewcommand{\arraystretch}{1.2}
\resizebox{\columnwidth}{!}{%
\begin{tabular}{l cc cc c}
\toprule
\textbf{Dataset} & \textbf{Video len.} & \textbf{$\ge 20$\,min}
 & \textbf{Temp.} & \textbf{De-leak} & \textbf{Neuro.} \\
\midrule
EndoVis18-VQA~\cite{allan20202018} & per-frame & --
 & \ding{55} & \ding{51} & \ding{55} \\
PitVQA~\cite{he2024pitvqa} & per-frame & --
 & \ding{55} & \ding{51} & \ding{51} \\
SurgViVQA~\cite{surgvivqa2025} & 8 frames & --
 & short & \ding{51} & \ding{55} \\
SVU-31K (test)~\cite{wang2025surgvidlm} & $13.9$~min & $28\%$
 & partial & \ding{55} & \ding{55} \\
\midrule
\textbf{MedClawBench (ours)} & $\mathbf{18.4}$~\textbf{min} & $\mathbf{41\%}$
 & \ding{51} & \ding{51} & \ding{51} \\
\quad\emph{neurosurgery split} & $28.5$~min & $71\%$
 & \ding{51} & \ding{51} & \ding{51} \\
\bottomrule
\end{tabular}%
}
\end{table}

\subsection{Data Construction}
\label{sec:data}
\begin{figure*}[t]
\centering
\includegraphics[width=0.94\textwidth]{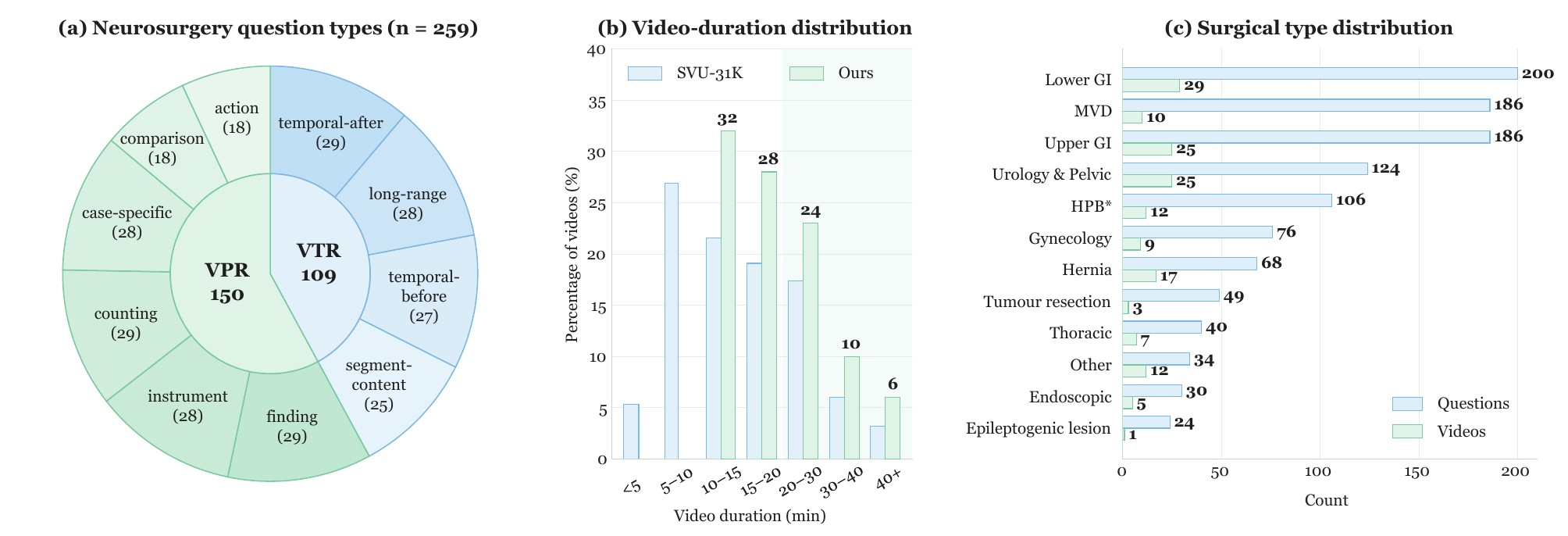}
\vspace{-5pt}
\caption{\textbf{MedClawBench statistics.}
\textbf{(a)} Neurosurgery question types ($259$): the inner ring is the Visual
Temporal (VTR, $109$) vs.\ Visual Perception (VPR, $150$) split and the outer ring
the ten fine-grained types.
\textbf{(b)} Source-video duration for MedClawBench (Ours) against SVU-31K, as the
percentage of videos per bin; our filter drops all sub-ten-minute videos and shifts
mass toward the long ($\ge 20$~min, shaded) tail.
\textbf{(c)} Surgical-type distribution over both splits (\# questions, \# videos),
with the neurosurgery half split into MVD, tumour resection and epileptogenic-lesion
resection (HPB* stands for Hepatobiliary \& Pancreas).}
\label{fig:benchmark_stats}
\vspace{-10pt}
\end{figure*}

MedClawBench contains $1{,}123$ questions drawn from two complementary sources,
so that it spans both a long, self-recorded surgical domain and a large,
externally sourced lecture domain. Fig.~\ref{fig:benchmark_stats} summarizes its
composition along three axes: question type (Fig.~\ref{fig:benchmark_stats}(a)),
source-video duration (Fig.~\ref{fig:benchmark_stats}(b)), and surgical type
(Fig.~\ref{fig:benchmark_stats}(c)).

\noindent\textbf{Neurosurgery.}
The first half comprises $259$ questions over 14 long neurosurgery
recordings from one centre, spanning three procedure types: microvascular
decompression (MVD, ten videos, $186$ questions), tumour resection (three videos,
$49$) and epileptogenic-lesion resection (one video, $24$). These procedures are
long (a $28.5$-minute median, with $71\%$ running beyond twenty minutes) and
visually homogeneous, making neurosurgery the harder, long-horizon half of the
benchmark. General-purpose vision models lack the clinical
knowledge to caption these procedures reliably, and automatically generated golds
were frequently ``correct textbook knowledge unrelated to the video,'' as flagged
by our neurosurgeon collaborator. We therefore \textbf{ground the labels in expert
neurosurgeon annotation} rather than model captions: a neurosurgeon segments each
procedure into time intervals and annotates the clinical event in each, and,
conditioned on these events, GPT-5.5 drafts diversified, event-anchored questions
whose gold is tied to a specific annotated interval, with embedding-based
de-duplication. Each candidate then passes the grounding filters and the faithfulness check below,
and is labelled with one of ten fine-grained types (Fig.~\ref{fig:benchmark_stats}(a))
that partition into VTR ($109$) and VPR ($150$), following Section~\ref{sec:task}.
The temporal (VTR) types ask what step immediately follows (\emph{temporal-after}) or
precedes (\emph{temporal-before}) a named event, how distant phases relate
(\emph{long-range-temporal}), or what unfolds within a stated interval
(\emph{segment-content}). The perceptual (VPR) types instead ask about a single
moment: which \emph{instrument} performs a step, what \emph{action} occurs, a
\emph{counting} of items, a \emph{finding} (an observed anatomical or pathological
detail), a \emph{comparison} (relating two
structures or states visible at that moment), or a \emph{case-specific} detail unique to the individual procedure.

\noindent\textbf{SVU (lecture).}
We use only the question--answer pairs from the test split of SVU-31K, a
lecture-video corpus released by SurgVidLM~\cite{wang2025surgvidlm}, and apply the
filtering cascade below to retain $864$ questions, reusing their original Visual
Temporal Reasoning ($163$) and Visual Perception Reasoning ($701$) labels.

\subsection{Filtering}
\label{sec:filter}
Every candidate must survive a cascade of filters that certify it can be answered
\emph{only} by watching the moment it targets. All candidates must pass the shared
duration, blind, and local filters; SVU questions additionally pass a narration
filter, and neurosurgery questions a faithfulness filter.
\textbf{(A) Duration:} we keep only questions whose source video is at least ten
minutes long, so that the benchmark stresses long-horizon reasoning over an extended
procedure rather than perception on a short clip.
For the answerability filters that follow we let a
restricted answerer respond and use an automatic judge (GPT-5.5) to score its
answer's overall correctness against the gold on a $1$--$5$ scale, against a fixed
threshold $\tau{=}3$.
\textbf{(B) Blind:} a text-only model (DeepSeek) answers with no video access,
yielding $s_B$, and we require $s_B<\tau$ to drop questions solvable from priors
alone.
\textbf{(C) Local:} a model answers using \emph{only} the dense observations
extracted from the targeted window, yielding $s_C$, and we require $s_C\ge\tau$ so
that a kept question is answerable from its own interval of visible evidence rather
than a coarse whole-procedure overview. 
\textbf{(D) Narration (SVU):} we re-answer each question from the audio track
alone, with no frames, and discard it if this scores $\ge \tau$, since its answer
would then be recoverable by listening rather than by watching.
\textbf{(E) Faithfulness (neurosurgery):} since the neurosurgery question and gold
are drafted by GPT-5.5, a fact-checker checks them against the neurosurgeon's
interval annotation and keeps a candidate only if it aligns with that annotation,
discarding any content GPT-5.5 hallucinated beyond it.

\section{Experiments}
\label{sec:experiment}

\subsection{Setup}
We evaluate on \textbf{MedClawBench}, comprising $259$ long neurosurgery
questions and $864$ held-out SVU lecture-video questions, for $1{,}123$ questions
in total. 
Every answer is scored on four dimensions, including correctness of information (CI), detail orientation
(DO), contextual understanding (CU), and temporal understanding
(TU)~\cite{wang2025surgvidlm}, each on a scale $1$--$5$ by an automatic judge, and we report
the average (Avg) of the four. Higher is better on all metrics. Our agent is
implemented on top of the open-source Hermes agent
framework.\footnote{\url{https://github.com/nousresearch/hermes-agent}}

\subsection{Baselines}
We compare against three families of models. \emph{General video-language
models}: Qwen2.5-VL-7B-Instruct~\cite{wu2025qwen},
InternVL3-8B~\cite{zhu2025internvl3}, and VideoLLaMA3-7B~\cite{zhang2025videollama}.
\emph{Video-reasoning models}: Video-R1-7B~\cite{feng2026video},
VideoChat-R1-7B~\cite{li2025videochat}, and Video-RFT~\cite{wang2026videorft}.
\emph{Long-video understanding agents}: LongVT-7B-RFT~\cite{yang2026longvt}
and ReWatch-R1-7B~\cite{zhang2025rewatch}. We further include the surgical
video-language model SurgVidLM~\cite{wang2025surgvidlm}. Finally, as the
one-shot vision backends inside our own harness, we report Gemini-3.1-flash-lite, Gemini-3.5-flash, GPT-5.5, and Opus-4.8 answering each question in a single
pass. \textbf{MedClaw} is our frozen-weight agent, which issues an auditable
sequence of tool calls (viewing, cropping, frame inspection, knowledge
retrieval), including a coarse-to-fine pass inside the full-video view and the
discovered directed re-look skill.

\begin{table*}[t]
\centering
\caption{Main results on MedClawBench, broken down by reasoning type
(\textbf{VTR}, \textbf{VPR}) with the four scoring dimensions each $1$--$5$: CI
(correctness), DO (detail), CU (context), TU (temporal). Best in \textbf{bold},
second best \underline{underlined}. Shaded cells mark the one-shot backbone used
for the \textbf{$\Delta$} row. MedClaw uses the flash visual backbone on neurosurgery and the
cheaper lite backbone on SVU, and the \textbf{$\Delta$} row compares against the corresponding
one-shot baseline in each split.}
\label{tab:main}
\footnotesize
\setlength{\tabcolsep}{2.6pt}
\renewcommand{\arraystretch}{1.15}
\newcommand{\hlcell}[1]{\cellcolor{black!6}#1}
\begin{tabular}{@{} l l cccc cccc c cccc cccc @{}}
\toprule
\multirow{3}{*}{\textbf{Cat.}} & \multirow{3}{*}{\textbf{Method}}
 & \multicolumn{8}{c}{\textbf{Neurosurgery ($259$)}} & \phantom{a}
 & \multicolumn{8}{c}{\textbf{SVU ($864$)}} \\
\cmidrule(lr){3-10} \cmidrule(lr){12-19}
 & & \multicolumn{4}{c}{VTR\,($109$)} & \multicolumn{4}{c}{VPR\,($150$)} & &
     \multicolumn{4}{c}{VTR\,($163$)} & \multicolumn{4}{c}{VPR\,($701$)} \\
\cmidrule(lr){3-6} \cmidrule(lr){7-10} \cmidrule(lr){12-15} \cmidrule(lr){16-19}
 & & CI & DO & CU & TU & CI & DO & CU & TU & &
     CI & DO & CU & TU & CI & DO & CU & TU \\
\midrule
\multirow{3}{*}{VLM}
 & Qwen2.5-VL-7B  & 1.62 & 1.31 & 2.28 & 2.01 & 1.80 & 1.61 & 2.53 & 2.90 & & 1.64 & 1.54 & 2.31 & 2.01 & 2.25 & 1.85 & 2.84 & 2.73 \\
 & InternVL3-8B   & 1.35 & 1.05 & 1.78 & 1.65 & 1.65 & 1.43 & 2.29 & 2.60 & & 1.51 & 1.18 & 1.98 & 1.75 & 1.81 & 1.16 & 2.11 & 2.09 \\
 & VideoLLaMA3-7B & 1.49 & 1.09 & 1.96 & 1.74 & 1.73 & 1.39 & 2.28 & 2.50 & & 1.60 & 1.29 & 2.17 & 1.79 & 1.87 & 1.34 & 2.26 & 2.28 \\
\midrule
\multirow{3}{*}{Reasoning}
 & Video-R1-7B     & 1.61 & 1.17 & 2.24 & 2.28 & 1.77 & 1.55 & 2.55 & 3.03 & & 1.67 & 1.31 & 2.33 & 2.02 & 2.10 & 1.50 & 2.68 & 3.04 \\
 & VideoChat-R1-7B & 1.75 & 1.29 & 2.33 & 2.03 & 2.04 & 1.65 & 2.67 & 3.02 & & 1.87 & 1.61 & 2.46 & 2.12 & 2.30 & 1.74 & 2.77 & 2.74 \\
 & Video-RFT       & 1.64 & 1.10 & 2.06 & 1.94 & 2.03 & 1.51 & 2.45 & 3.01 & & 1.85 & 1.42 & 2.37 & 2.06 & 2.22 & 1.64 & 2.64 & 2.76 \\
\midrule
\multirow{2}{*}{Agent}
 & LongVT-7B-RFT   & 1.75 & 1.26 & 2.18 & 1.93 & 1.80 & 1.40 & 2.23 & 2.29 & & 1.53 & 1.24 & 2.04 & 1.65 & 1.98 & 1.43 & 2.47 & 2.45 \\
 & ReWatch-R1-7B   & 1.46 & 1.10 & 1.79 & 1.53 & 1.57 & 1.16 & 1.79 & 1.85 & & 1.70 & 1.23 & 2.18 & 1.80 & 1.91 & 1.30 & 2.35 & 2.29 \\
\midrule
Surgical
 & SurgVidLM       & 1.53 & 1.28 & 2.21 & 1.83 & 1.66 & 1.55 & 2.53 & 2.63 & & 1.82 & 1.67 & 2.61 & 1.97 & 2.24 & 1.90 & 2.87 & 2.69 \\
\midrule
1-shot
 & Gemini-3.1-flash-lite & 2.10 & 1.80 & 2.83 & \underline{2.59} & 2.27 & 2.12 & 3.07 & 3.35 & & \hlcell{2.60} & \hlcell{2.53} & \hlcell{3.28} & \hlcell{2.77} & \hlcell{3.40} & \hlcell{3.14} & \hlcell{3.88} & \hlcell{3.73} \\
 & Gemini-3.5-flash & \hlcell{1.96} & \hlcell{1.77} & \hlcell{2.71} & \hlcell{2.28} & \hlcell{2.23} & \hlcell{2.13} & \hlcell{2.96} & \hlcell{3.36} & & \underline{2.79} & \underline{2.77} & \underline{3.44} & \underline{2.98} & \underline{3.62} & \underline{3.39} & \underline{4.05} & \underline{3.94} \\
 & Opus-4.8         & \underline{2.15} & \underline{2.01} & \underline{2.93} & 2.49 & \underline{2.29} & \underline{2.40} & \underline{3.19} & \underline{3.43} & & 2.36 & 2.34 & 3.13 & 2.52 & 2.57 & 2.42 & 3.15 & 2.96 \\
\midrule
\multirow{2}{*}{\textbf{Ours}}
 & \textbf{MedClaw} & \textbf{2.40} & \textbf{2.28} & \textbf{3.19} & \textbf{2.80} & \textbf{2.64} & \textbf{2.69} & \textbf{3.38} & \textbf{3.53} & & \textbf{3.03} & \textbf{3.09} & \textbf{3.69} & \textbf{3.18} & \textbf{3.67} & \textbf{3.65} & \textbf{4.17} & \textbf{3.95} \\
 & \textbf{$\Delta$ (1-shot baseline)} & \textcolor{green!50!black}{+0.44} & \textcolor{green!50!black}{+0.51} & \textcolor{green!50!black}{+0.48} & \textcolor{green!50!black}{+0.52} & \textcolor{green!50!black}{+0.41} & \textcolor{green!50!black}{+0.56} & \textcolor{green!50!black}{+0.42} & \textcolor{green!50!black}{+0.17} & & \textcolor{green!50!black}{+0.43} & \textcolor{green!50!black}{+0.56} & \textcolor{green!50!black}{+0.41} & \textcolor{green!50!black}{+0.41} & \textcolor{green!50!black}{+0.27} & \textcolor{green!50!black}{+0.51} & \textcolor{green!50!black}{+0.29} & \textcolor{green!50!black}{+0.22} \\
\bottomrule
\end{tabular}
\end{table*}

\subsection{Main Results}
Table~\ref{tab:main} reports results separately for Visual Temporal Reasoning
(VTR) and Visual Perception Reasoning (VPR), which together make up each split.
On the long neurosurgery split, MedClaw uses the stronger flash visual
backbone and reaches a four-dimensional average of $2.90$, ahead of the strongest
open-source video model we test (VideoChat-R1-7B at $2.14$) by $+0.76$. On SVU,
where the clips are shorter and the two visual backbones are comparable, MedClaw
uses the cheaper lite visual backbone. The table therefore reports a single
MedClaw row with the best deployed configuration for each split: flash for
neurosurgery and lite for SVU. The \emph{$\Delta$} row compares MedClaw against the
corresponding one-shot backbone in each split (Gemini-3.5-flash for neurosurgery
and Gemini-3.1-flash-lite for SVU): using the same visual backbone, MedClaw
improves every dimension by $+0.17$ to $+0.56$, which shows that our agent
harness is what drives the gain.
Every open-source baseline, including the surgical video-language model
SurgVidLM ($1.94$) and the
re-watching model ReWatch-R1-7B ($1.54$), trails all of the one-shot backends on
neurosurgery, underscoring how far long-horizon neurosurgery falls
outside their training distribution.

\subsection{Ablation Studies}
We ablate using the long neurosurgery split ($259$ questions).

\noindent\textbf{The harness lifts diverse backbones.}
We compare two settings for each backbone (Table~\ref{tab:backbone_lift}): (\romannumeral1) a one-shot
pass, where the model answers the question directly from the video, against
(\romannumeral2) wrapping the \emph{same} model inside our harness as the visual sub-agent that
sees the video, in place of Gemini. Across every backbone we test, wrapping it in
our harness improves its four-dimensional average, by $+0.54$ for Qwen2.5-VL,
$+0.55$ for Video-R1, and up to $+1.02$ for ReWatch-R1-7B, indicating that our generic harness delivers consistent gains when applied to a broad range of video-understanding models.
%

\begin{table}[!htbp]
\centering
\caption{Backbone-lift ablation on the neurosurgery split ($259$ questions):
one-shot vs.\ the same backbone wrapped as the sub-agent in our frozen-weight
harness. Scores are the average over the four dimensions. Best per column in
\textbf{bold}.}
\label{tab:backbone_lift}
\footnotesize
\setlength{\tabcolsep}{8pt}
\renewcommand{\arraystretch}{1.3}
\begin{tabular}{l ccc}
\toprule
\textbf{Setting} & \textbf{Video-R1-7B} & \textbf{Qwen2.5-VL-7B} & \textbf{ReWatch-R1-7B} \\
\midrule
one-shot           & 2.06 & 2.04 & 1.54 \\
\textbf{+ harness} & \textbf{2.60} & \textbf{2.58} & \textbf{2.56} \\
\midrule
\textit{Gain}      & \textcolor{green!50!black}{+0.55} & \textcolor{green!50!black}{+0.54} & \textcolor{green!50!black}{+1.02} \\
\bottomrule
\end{tabular}
\end{table}

\noindent\textbf{Removing each component.}
Starting from the full model, we remove or alter one learned component at a time
(Table~\ref{tab:component_ablation}): the discovered \emph{directed re-look} skill,
\emph{coarse-to-fine} localization, and \emph{Heuristic Skill Distillation} itself (replacing
the distilled skills with a hand-written general looking-discipline skill). Directed
re-look matters most: removing it drops the average by $-0.38$. This skill targets
\emph{before} and \emph{after} temporal questions, where after localizing the named
event it re-examines the window adjacent to that event, since the answer lies just
outside the moment the event occupies rather than in the coarse first pass.
Coarse-to-fine localization ($-0.16$) shows that narrowing from the whole procedure
down to the exact moment, rather than reading one compressed pass, is needed to
find the evidence a question targets. Heuristic Skill Distillation ($-0.13$) shows
that skills mined and reward-gated from the agent's own traces outperform a
hand-written looking-discipline skill, so the distilled skills contribute beyond
generic looking advice. Every component thus makes a measurable contribution, and
directed re-look is the largest.

\begin{table}[!htbp]
\centering
\caption{Component ablation on the neurosurgery split ($259$ questions): from the
full MedClaw harness, we remove or alter one learned component at a time. Scores
are the four dimensions and their average; $\Delta$ is the change in average from
the full model.}
\label{tab:component_ablation}
\footnotesize
\setlength{\tabcolsep}{4.5pt}
\renewcommand{\arraystretch}{1.2}
\begin{tabular}{l cccc c c}
\toprule
\textbf{Configuration} & CI & DO & CU & TU & \textbf{Avg} & \textbf{$\Delta$} \\
\midrule
\textbf{MedClaw (full)}        & \nv{2.54} & \nv{2.52} & \nv{3.30} & \nv{3.22} & \nv{\textbf{2.90}} & --- \\
\midrule
w/o coarse-to-fine             & \nv{2.36} & \nv{2.36} & \nv{3.19} & \nv{3.03} & \nv{2.74} & \textcolor{red}{$-0.16$} \\
w/o directed re-look           & \nv{2.19} & \nv{2.17} & \nv{2.93} & \nv{2.76} & \nv{2.51} & \textcolor{red}{$-0.38$} \\
w/o HSD                        & \nv{2.37} & \nv{2.39} & \nv{3.17} & \nv{3.14} & \nv{2.77} & \textcolor{red}{$-0.13$} \\

\bottomrule
\end{tabular}
\end{table}

\noindent\textbf{Swapping the orchestrator.}
The harness should not depend on one particular reasoner. We keep the perception side
fixed and swap only the text-only orchestrator (Table~\ref{tab:orch}). Replacing
Opus-4.8 with either open-weight DeepSeek reasoner costs about a third of a point
(DeepSeek-V3 $-0.33$, DeepSeek-R1 $-0.34$). A
weaker orchestrator thus degrades the harness gracefully rather than breaking it,
while Opus-4.8 remains the strongest choice for driving the tool-call loop.

\begin{table}[!htbp]
\centering
\caption{Orchestrator ablation on the neurosurgery split ($259$ questions): the
visual sub-agent and all tools are held fixed; only the text-only orchestrator
changes. Scores are the four dimensions and their average; $\Delta$ is the change
from the Opus-4.8 orchestrator.}
\label{tab:orch}
\footnotesize
\setlength{\tabcolsep}{4.5pt}
\renewcommand{\arraystretch}{1.2}
\begin{tabular}{l cccc c c}
\toprule
\textbf{Orchestrator} & CI & DO & CU & TU & \textbf{Avg} & \textbf{$\Delta$} \\
\midrule
\textbf{Opus-4.8}              & \nv{2.54} & \nv{2.52} & \nv{3.30} & \nv{3.22} & \nv{\textbf{2.90}} & --- \\
\midrule
DeepSeek-V3                    & \nv{2.07} & \nv{2.36} & \nv{2.94} & \nv{2.89} & \nv{2.57} & \textcolor{red}{$-0.33$} \\
DeepSeek-R1                    & \nv{2.22} & \nv{2.21} & \nv{3.04} & \nv{2.75} & \nv{2.56} & \textcolor{red}{$-0.34$} \\
\bottomrule
\end{tabular}
\end{table}

\begin{table}[!htbp]
\centering
\caption{Judge reliability: quadratic weighted Cohen's $\kappa$ between the
GPT-5.5 judge and expert ratings, per dimension, on each split.}
\label{tab:kappa}
\setlength{\tabcolsep}{8pt}
\renewcommand{\arraystretch}{1.25}
\begin{tabular}{lcc}
\toprule
\multirow{2}{*}{\textbf{Dimension}}
 & \textbf{Neurosurgery} & \textbf{SVU} \\
 & $\kappa$ (GPT-5.5) & $\kappa$ (GPT-5.5) \\
\midrule
CI (correctness)        & \nv{0.71} & \nv{0.87} \\
DO (detail)             & \nv{0.66} & \nv{0.86} \\
CU (context)            & \nv{0.72} & \nv{0.85} \\
TU (temporal)           & \nv{0.74} & \nv{0.86} \\
\midrule
Overall                 & \nv{0.72} & \nv{0.86} \\
\bottomrule
\end{tabular}
\end{table}

\noindent\textbf{Judge reliability.}
To confirm that the automatic four-dimensional scores reflect clinical judgment
rather than a self-referential metric, we validate our GPT-5.5 judge against
neurosurgeon ratings, reporting the quadratic weighted Cohen's
$\kappa$~\cite{cohen} between judge and expert per dimension on a held-out subset
(Table~\ref{tab:kappa}). On the neurosurgery subset ($30$ items scored by a
neurosurgeon), the judge reaches substantial agreement on every dimension
(overall $\kappa = 0.72$), and its mean scores track the expert's almost exactly
($2.95$ vs.\ $2.94$ overall). On the SVU subset ($70$ items scored by surgical
experts), agreement is almost perfect on every dimension (overall $\kappa = 0.86$,
mean $3.64$ vs.\ $3.65$); the higher figure is expected, since the lecture domain
is less specialized than neurosurgery and thus easier to rate consistently. Across
both splits the automatic scores align closely with expert judgment.

\section{Limitations and Future Work}
Our benchmark only contains a limited number of questions,
and the neurosurgery portion, in particular, is drawn from only a small set of
recordings. In future work, we plan to annotate a larger
collection of neurosurgery videos, focusing on long recordings beyond twenty
minutes, and to generate more temporal-reasoning questions over them, so as to
build a substantially larger benchmark on which to further test the performance
and generality of our method.

\section{Conclusion}
We addressed long-horizon temporal reasoning on long surgical videos
without training any model to decide where to look. Separating reasoning from
perception, we built a training-free agent harness in which a text orchestrator
plans and unmodified vision--language sub-agents look, and a gradient-free,
reward-gated Heuristic Skill Distillation loop mines reusable retrieval skills, notably
directed re-look, from about one hundred preference examples rather than by
fine-tuning. On MedClawBench, our de-leaked and doctor-grounded benchmark, the
harness surpasses one-shot VLMs and general video-agent
frameworks across all four dimensions, with the largest margins on the long, out-of-domain
neurosurgery videos where long-horizon temporal reasoning is most stressed. Our
ablations show that directed re-look contributes the most, that the harness
delivers consistent gains across diverse frozen backbones, and that it survives
swapping the orchestrator. Two properties of this training-free formulation are of particular relevance to
clinical practice: auditability and data-efficient adaptation. Because every
retrieval step is realized as an explicit tool call over unchanged weights, the
evidence underlying each answer is exposed as an inspectable trace, allowing a
clinician to review where the agent attended and how its conclusion was derived.
Adaptation, in turn, is confined to an external skill library rather than to the
model parameters, enabling extension to unseen procedures or specialties from a
small number of examples, without retraining or large-scale annotation.

\section*{References}

\bibliographystyle{IEEEtran}
\bibliography{main}

\end{document}